\documentclass[conference]{IEEEtran}

\PassOptionsToPackage{protrusion, expansion}{microtype} 

\usepackage{amsmath, amsthm}
    
\usepackage{graphicx, xcolor} 
\usepackage{hyperref, url}
\usepackage{array}
\usepackage{booktabs, multirow}
\usepackage{caption}
\usepackage{subcaption}
\usepackage{enumitem}
\usepackage{setspace}

\usepackage[ruled, linesnumbered, norelsize]{algorithm2e}

\begin{document}

\title{Enhancing Lossy Compression Through Cross-Field Information for Scientific Applications}

\makeatletter
\newcommand{\linebreakand}{%
  \end{@IEEEauthorhalign}
  \hfill\mbox{}\par
  \mbox{}\hfill\begin{@IEEEauthorhalign}
}
\makeatother

\author{
\centering
\IEEEauthorblockN{Youyuan Liu}
\IEEEauthorblockA{\textit{Temple University}\\
Philadelphia, PA, USA \\
youyuan.liu@temple.edu}
\and
\IEEEauthorblockN{Wenqi Jia}
\IEEEauthorblockA{\textit{University of Texas at Arlington}\\
Arlington, TX, USA \\
wxj1489@mavs.uta.edu}
\and
\IEEEauthorblockN{Taolue Yang}
\IEEEauthorblockA{\textit{Temple University}\\
Philadelphia, PA, USA \\
taolue.yang@temple.edu}
\linebreakand
\IEEEauthorblockN{Miao Yin}
\IEEEauthorblockA{\textit{University of Texas at Arlington}\\
Arlington, TX, USA \\
miao.yin@uta.edu}
\and
\IEEEauthorblockN{Sian Jin\thanks{Corresponding author: Sian Jin (\url{sian.jin@temple.edu}), College of Science and Technology, Temple University, Philadelphia, PA, USA.}}
\IEEEauthorblockA{\textit{Temple University}\\
Philadelphia, PA, USA \\
sian.jin@temple.edu}
}

\maketitle

\begin{abstract}
Lossy compression is one of the most effective methods for reducing the size of scientific data containing multiple data fields.
It reduces information density through prediction or transformation techniques to compress the data. 
Previous approaches use local information from a single target field when predicting target data points, limiting their potential to achieve higher compression ratios.
In this paper, we identified significant cross-field correlations within scientific datasets. We propose a novel hybrid prediction model that utilizes CNN to extract cross-field information and combine it with existing local field information. Our solution enhances the prediction accuracy of lossy compressors, leading to improved compression ratios without compromising data quality. We evaluate our solution on three scientific datasets, demonstrating its ability to improve compression ratios by up to 25\% under specific error bounds. Additionally, our solution preserves more data details and reduces artifacts compared to baseline approaches. 
\end{abstract}


\pagestyle{plain}
\setlength{\textfloatsep}{6pt}
\section{Introduction}
\label{sec:introduction}

Large-scale scientific simulation now play an important role in recent science research. These simulations can easily generate large amount of data. Take Nyx~\cite{nyx} as an example, one Nyx cosmological simulation with a resolution of 4096 × 4096 × 4096 cells can generate up to 2.8 TB of data for a single snapshot; a total of 2.8 PB of storage is needed, assuming the simulation runs 5 times with 200 snapshots dumped per simulation. The large volume of data presents two challenges: First, storing this scale of data entirely on disk is difficult, even for supercomputers. Second, the large data volume can consume significant time during file I/O or data transfer between devices, due to limitations in I/O bandwidth.

Lossy compression is one of the best solutions to address this issue, as it can efficiently reduce the data size while only introducing data distortion within a controllable range. Compared to lossless compression, this approach is particularly effective for scientific data with significantly high compression ratio. 
The new generation of lossy compressors for scientific data, including SZ~\cite{di2016fast, tao2017significantly, sz18, sz3} and ZFP~\cite{zfp}, have been widely employed in various systems and frameworks to reduce data sizes and enhance performance~\cite{eisenman2018reducing, jin2019deepsz, jin2021comet, wang2023amric}.
Researchers have also conducted extensive studies around various lossy compressors to improve their performance. 
For example, GPU variants of SZ and ZFP have been proposed to achieve exceptional throughput~\cite{tian2020cusz, cuZFP}, while novel predictors have been introduced to enhance compression ratios~\cite{sz3,liu2021exploring}.

Recently, researchers have tried to use machine learning to improve the performance of lossy compression. 
For instance, Liu et al. introduced AE-SZ~\cite{liu2021exploring}, a framwork which use a convolutional autoencoder to improve error-bounded lossy compression for scientific data, and this framework exhibits better rate-distortion performance than SZ. 
However, several challenges remain: (1) The model size of the autoencoder is often quite large; for example, AE-SZ uses a model with approximately 380,000 parameters to predict the CLDTOT field in the CESM-ATM dataset, which contains only 1800 × 3600 data points. (2) Training the model can be very time-consuming, taking over 20 hours in some extreme cases.
Jia et al. proposed GWLZ~\cite{jia2024gwlzgroupwiselearningbasedlossy}, a learning-based lossy compression framework, which uses multiple DNN models to reduce the reconstruction error introduced by SZ. 

However, existing studies only use information from the target field itself for prediction~\cite{sz3}, transformation~\cite{zfp}, or enhancement~\cite{jia2024gwlzgroupwiselearningbasedlossy}. We have observed that there are strong correlations between different fields within the same dataset. In Figure \ref{fig:fig-compare}, we present the visualization results of the 49th slice along the first dimension of the u, v, and w fields (i.e., zonal, meridional, and vertical wind speeds respectively) from the SCALE dataset. It is clear that they share similar structures, though their distributions in finer details are not the same, and their value ranges differ as well.

\label{sec:3.1}
\begin{figure}[]
    \centering
   \includegraphics[width=0.95\linewidth]{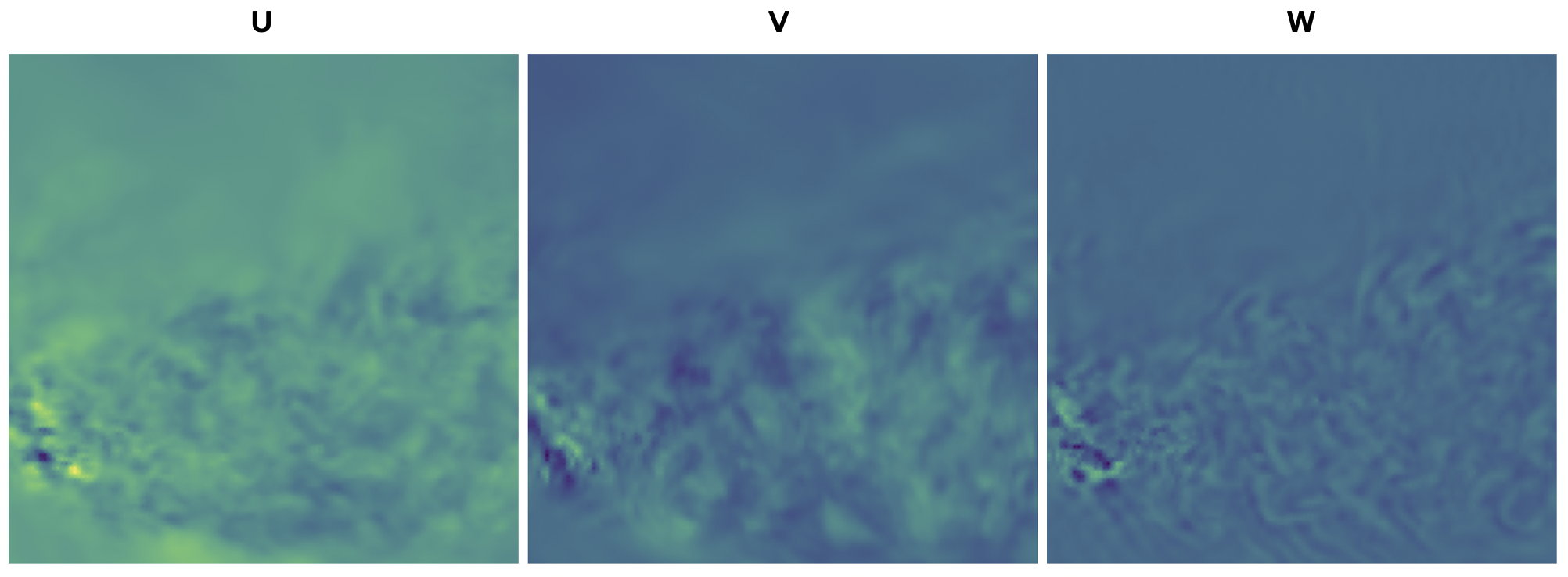}
    \caption{The visualization of the 49th slice (along the first dimension) of U, V, and W field of SCALE dataset. Dimension in 1200$\times$1200. A distinct yet nonlinear correlation between data fields can be observed.}
    \label{fig:fig-compare}
\end{figure}

Existing compressors for scientific data do not take advantage of this cross-field information. Therefore, in this paper, we propose a cross-field predictor to capture this information. We first use a CNN to predict the first-order backward difference of the target field by inputting the first-order backward difference of other fields (which we refer to as anchor fields). Then, we combine this difference predictor with the original prediction-based compressor’s predictor using a hybrid network to achieve a better prediction result and ultimately improve the compression ratio. To the best of our knowledge, this is the first approach that integrates cross-field information into a compression method. In summary, the contributions of our solution are as follows:

\begin{itemize}[topsep=2pt,leftmargin=1em]
    \item We observe and identify significant cross-field information between different fields within the same dataset, which has been largely overlooked in traditional compression methods.
    \item We propose a CNN-based model specifically designed to extract and leverage this cross-field information, enhancing the prediction accuracy.
    \item We develop a compact model that effectively hybrid cross-field information with local information, improving the overall prediction and compression performance.
    \item We evaluate of our solution to show its effectiveness across various datasets. Our solution can significantly improve the compression ratio by up to 20\% under certain error bounds, and it is applicable across multiple datasets.
\end{itemize}

In Section 2, we discuss the background of the research. Section 3 details our proposed solution, which utilizes cross-field information to enhance the performance of prediction-based lossy compressors. In Section 4, we present the results of our evaluation across a range of datasets and provide a comprehensive analysis of the findings. Finally, in Section 5, we conclude the paper and discuss potential directions for future work.

\section{Background And Motivation}
\label{sec:background}
\subsection{Lossy Compression}

In recent years, a new generation of high-accuracy lossy compressors for scientific data has been proposed and developed for scientific floating-point data, such as SZ~\cite{di2016fast, tao2017significantly, sz18} and ZFP~\cite{zfp}.
They have been widely employed in various systems and frameworks to reduce data sizes and enhance performance~\cite{eisenman2018reducing, jin2019deepsz, jin2021comet, wang2023amric}.
Compared to lossless compression, lossy compression can provide significantly higher compression ratio by losing non-critical information.
Unlike traditional lossy compressors such as JPEG~\cite{wallace1992jpeg} 
which are designed for images (in integers), SZ and ZFP are designed to compress floating-point data and can provide a strict error-controlling scheme based on user's requirements.
Generally, lossy compressors provide multiple compression modes, such as the error-bounding mode. 
The error-bounding mode requires users to set an error type (e.g., relative error bound) and a bound value (e.g., $10^{-3}$). The compressor ensures that differences between original and reconstructed data do not exceed the error bound.

Specifically, prediction-based lossy compression, such as SZ, involves three main steps:
(1) Each data point's value is predicted based on its neighboring points, using an adaptive, best-fit prediction method.
(2) The difference between the actual value and the predicted value is quantized, based on the user-set compression mode and error bound.
(3) Customized Huffman coding and additional lossless compression are applied to achieve a high compression ratio.
The two most important metric types to evaluate the performance of lossy compression are: (1) compression ratio, i.e., the ratio between original data size and compressed data size, or bit-rate, i.e., the number of bits on average for each data point (e.g., 32/64 for single/double-precision floating-point data) before compression; and (2) data distortion metrics such as Peak Signal-to-Noise Ratio (PSNR) and Structural Similarity Index (SSIM) to measure the reconstructed data quality compared to the original data.
Previous studies have focused on using surrounding data values to predict the current value, including using predictors such as Lorenzo~\cite{sz16}, Regression~\cite{sz17}, Interpolation~\cite{sz3}, and autoencoder~\cite{liu2021exploring}. 
However, cross-field information has largely been overlooked.

\subsection{Convolutional Neural Network}
Convolutional Neural Networks (CNNs) are a class of feedforward neural networks that use filters to learn data features. In a CNN, there are input layers, hidden layers, and output layers, with the hidden layers containing one or more convolutional layers. These convolutional layers learn hidden features in the input data by applying filters that slide across the data, eventually producing a feature map for the next layer to learn from. In addition to convolutional layers, CNNs often include fully connected layers, various pooling layers, activation layers such as ReLU, and other specialized layers. Due to their exceptional feature extraction capabilities, CNNs have seen successful applications in many fields, including image and video recognition~\cite{ivr}, image classification~\cite{cnn}, and image segmentation~\cite{imgseg}.

Additionally, CNNs have shown strong performance in prediction tasks~\cite{prediction}. The goal of this paper is to learn cross-field information, and the local distribution similarity is an important prior knowledge. CNNs are particularly effective at extracting local features~\cite{local_feature}, making them well-suited for this task. Furthermore, the Lorenzo predictor used in SZ can actually be viewed as a masked CNN with fixed parameters. Therefore, we aim to leverage CNNs for cross-field prediction.

\section{Methodology}
\label{sec:3.0}

In this section, we propose a machine learning-based method for predicting the current field using information from other fields. This additional information is utilized to enhance the existing Lorenzo predictor, with the objective of improving compression ratio. Our solution demonstrates significant improvements in compression ratio across various datasets and error bounds. By leveraging the mutual information between fields, our solution addresses several limitations of the traditional Lorenzo predictor and provides a more robust approach to data compression. The overview of our solution is shown in Figure \ref{fig:fig-overview}. When compressing the target field, we use anchor fields to perform cross-field prediction, which enhances the accuracy of the prediction process. We start by calculating the first-order backward difference of the anchor field. This difference is then input into a CNN model to predict the first-order backward differences of the target field across different dimensions. The dashed line in the figure indicates that the target field is only involved during the training phase of the CNN model. The results of the cross-field prediction are then combined with the traditional prediction using local field information through a hybrid prediction model, leading to a more accurate overall prediction. Finally, the prediction is processed through encoding and lossless compression, resulting in the compressed data.

\begin{figure}[]
    \centering
   \includegraphics[width=0.95\linewidth]{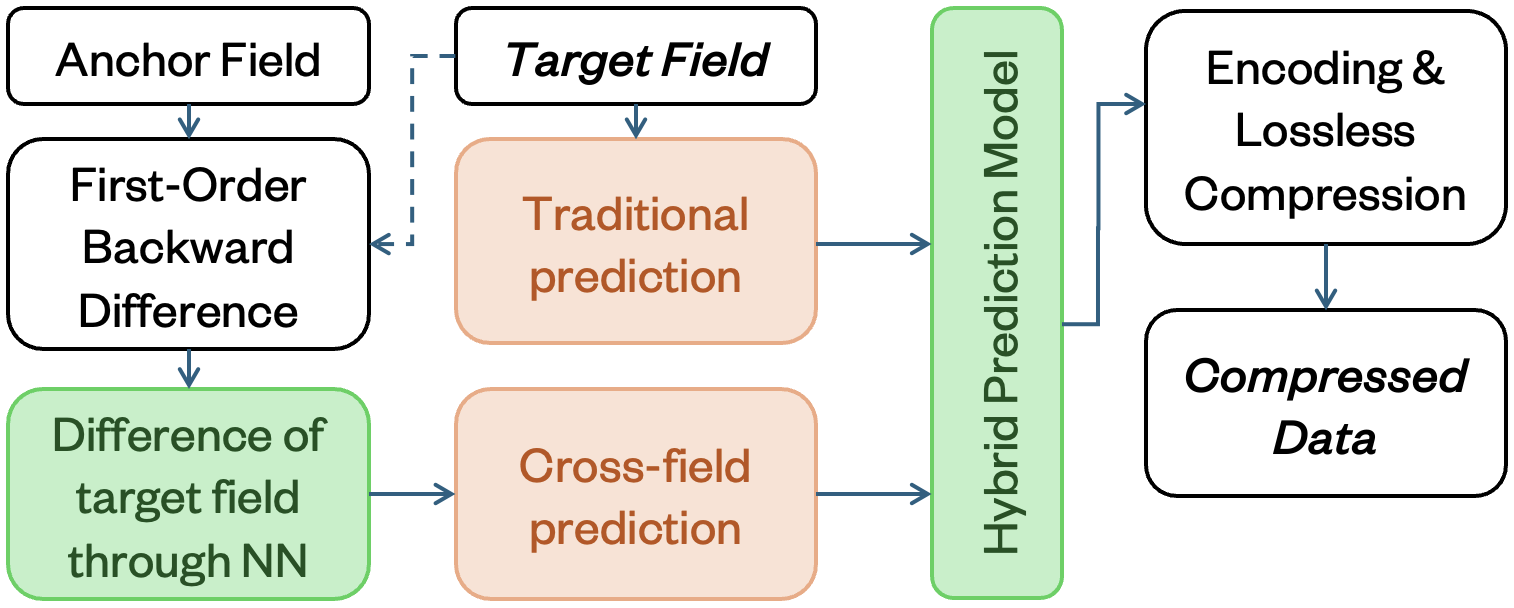}
    \caption{Overview of our proposed solution: enhancing the prediction effectiveness of lossy compressor by leveraging the mutual information between anchor fields and the target field, thereby improving the compression ratio.}
    \label{fig:fig-overview}
\end{figure}

\subsection{Cross-field Information}
In scientific datasets, there are often inherent physical or mathematical relationships between different fields. 
Traditional lossy compressors typically rely solely on information from the target field for compression (i.e., local field information), overlooking the valuable cross-field information that could be utilized to improve the compression ratio. 
For example, in the CESM-ATM dataset, the FLUT field closely mirrors the FLNT field, and the difference between the FLUTC and LWCF fields is also similar to the FLNT field in both value range and trend.
Meanwhile, not all cross-field knowledge is as straightforward as this example. 
In many cases, the relationships between related fields are far more complex. 
Nonetheless, when visualized, these fields often exhibit significant correlations, particularly in the area of extreme values. 
Effectively capturing this cross-field information could greatly enhance the accuracy of predictions. 
However, the complexity of these relationships presents challenges for traditional algorithms to extract them. 
Furthermore, the large number of fields introduces additional difficulties in applying conventional methods. 
Consequently, we have chosen to apply machine learning to address this task, as it is more adept at extracting complex, non-linear relationships.

\subsection{Cross-field Prediction}
Physical and mathematical relationships exist between different data fields. 
A straightforward approach to capture this information is to use a neural network to directly predict the values of the target field based on the data from anchor fields. 
However, our evaluation shows that this approach rarely performs well, often resulting in poor prediction accuracy. 
This is because: (1) The value ranges of the data fields can be large, leading to a loss of precision when machine learning models attempt to predict these values; (2) The value itself can be highly irregular and difficult to learn; and (3) Machine learning models require a large input area to capture the basic value range of the target values, which results in slow inference throughput and large model sizes.

To solve this problem, we use a method that learns to predict the first-order backward difference of the target field in all directions.
We propose a network named Cross-Field Neural Network (CFNN) to learn this backward difference. 
Compared to the original value, the first-order differences reflect local changes and are usually smoother, making them easier to learn with smaller input data area.
The value range of these differences is usually smaller, which helps with normalization. 
We choose to use the first-order differences of the original data from the anchor fields as the training data and use the \textbf{decompressed data} to predict the first-order differences of the target field. Although using decompressed data as training data can sometimes yield better prediction results, using the original data allows the CFNN to learn the true relationships between these fields. Additionally, it significantly reduces training time because, if we use decompressed data, the CFNN model would need to be retrained for each error bound.
In fact, we found that learning the first-order central difference works better than backward difference. 
However, considering our overall design, we chose to learn the first-order backward difference in this paper.
This choice will be explained in Section~\ref{sec:3.3}.

\subsection{Combine with SZ}
\label{sec:3.3}
In this paper, we aim to integrate the cross-field prediction to existing prediction-based lossy compressor.
Specifically, we use SZ~\cite{di2016fast, tao2017significantly, sz18, sz3} error-bounded lossy compression framework as baseline.
SZ supports several types of predictors: Lorenzo, regression, and interpolation—as well as their combinations. 
In this paper, we focus on enhancing with the Lorenzo predictor because (1) it is widely used and implemented in various SZ variants (e.g., cuSZ~\cite{tian2020cusz}, SZx~\cite{yu2022ultrafast}, cuSZp~\cite{huang2023cuszp}); and (2) it performs well across most datasets, particularly in scenarios involving typical compression ratios (i.e., 5–32).

In 2D datasets, predicting the data at position $(i,j)$ using the 1-layer Lorenzo predictor is given by $f_{Lorenzo}(i,j)=V(i-1,j)+V(i,j-1)-V(i-1,j-1)$, where $V$ represents the predicted value. 
Use this as a basis for reference., we formulate the first-order difference for predicting the data at position $(i,j)$ by $$
f_{cross-field-x}(i,j)= 
f(i-1,j)+d_x(i,j) $$
$$
f_{cross-field-y}(i,j)=
f(i,j-1)+d_y(i,j)
$$ where $d_x$ and $d_y$ are the backward differences predicted by cross-field information.
Our goal is to use CFNN to efficiently predict the $d_x(i,j)$ and $d_y(i,j)$ using information from anchor fields.
Note that while the Lorenzo predictor is similar in structure, it relies only on local information. 
A similar formulation is applied to 3D datasets.

\textbf{Hybrid prediction} refers to the process of combining multiple predicted values (i.e., $f_{Lorenzo}(i,j)$ and $f_{cross-field}(i,j)$) using a fast neural network to obtain a more accurate prediction, namely the hybrid prediction model. 
After cross-field and Lorenzo prediction, we have 
$n+1$ predicted values for an $n$-dimensional dataset. This includes $n$ values predicted by $n$-dimensional differences and the value predicted by the original Lorenzo predictor. 
Our proposed hybrid prediction model uses a neural network to achieve a more precise combination of the $n+1$ predicted values.

\begin{figure}[]
    \centering
   \includegraphics[width=0.95\linewidth]{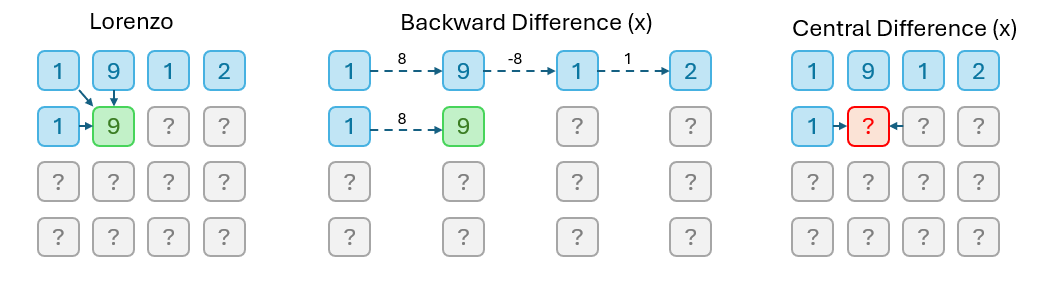}
    \caption{Both Lorenzo and backward difference predictor would only use data points which have already be predicted, while central difference predictor would use unpredicted data points.}
    \label{fig:fig-predictors}
\end{figure}

Our optimization focuses on enhancing the prediction stage. However, the choice of predictor must also consider the implications for subsequent steps.
Although our proposed cross field predictor only takes anchor fields as input and have no dependency within target field, the predicted first-order backward difference (e.g., $d_x(i,j)$) require previous values of target field to make the final value-prediction (e.g., $f_{cross-field}(i,j)$).
Figure \ref{fig:fig-predictors} shows that the cross-field predictor uses the same decompressed values of the target field as the Lorenzo predictor when predicting the current data point. 
As a result, there are no conflicts when running Lorenzo prediction and cross-field prediction together, while other predictors may have dependency conflicts with Lorenzo predictor. 
For example, the central difference predictor, shown in Figure ~\ref{fig:fig-predictors}, relies on data points on both sides of a given point while Lorenzo only depends data points on one side. The decompression order for this predictor differs from that of the Lorenzo predictor, making it incompatible with the Lorenzo predictor.

By combining cross-field information and local field information effectively, we can significantly improve the performance of our solution.

\subsection{Detailed Design}
\subsubsection{Use Dual-quant to solve RAW dependency}
In the SZ compressor, one critical challenge during the prediction and quantization stages is handling read-after-write (RAW) dependencies. In the original SZ algorithm, after predicting a data point, the error between the predicted and actual values is immediately quantized and written back to the dataset. This ensures consistency between compression and decompression, but it also introduces a RAW dependency, requiring each point to be processed sequentially. This sequential processing creates a significant performance bottleneck.

To address this issue, we implement a dual quantization approach, which is initially proposed in cuSZ~\cite{tian2020cusz}. Dual quantization involves two main steps: prequantization and postquantization. Here’s how it works:
\begin{itemize}
    \item \textbf{Prequantization}: The original dataset is first quantized based on a user-defined error bound (eb). This step converts the original values into prequantized values that are easier to manage and ensures they are multiples of the error bound. This step is computationally straightforward and avoids the RAW dependency because it does not rely on the predicted values from previous steps.
    \item \textbf{Postquantization}: After prequantization, prediction is performed using these prequantized values. The difference between the predicted value and the prequantized value is then calculated and stored. This difference, known as the postquantized value, does not introduce new errors and allows the prediction step to proceed without the sequential dependency that plagues the original SZ3 method.
\end{itemize}
By using dual quantization, we can eliminate the RAW dependency in compression process, enabling parallel processing of data points and significantly improving the compression throughput. During decompression, the prequantized values are used to reconstruct the original data by adding the postquantized differences, ensuring accuracy and consistency with the compressed data.
\subsubsection{Network For Cross-field Prediction}
The CFNN is a specialized network designed for scientific data prediction, combining Depthwise Separable Convolutions~\cite{Chollet_2017_CVPR} and a Channel Attention mechanism~\cite{Woo_2018_ECCV} to enhance computational efficiency and prediction accuracy. It takes the first-order backward differences of the anchor fields as input and outputs the predicted first-order backward differences of the target field. The network architecture of CFNN is illustrated in Figure \ref{fig:fig-cfnn}. The network begins with an initial convolutional layer that performs preliminary feature extraction from the input data, capturing local spatial information. Following this, the Depthwise Separable Convolution module operates, consisting of a Depthwise Convolution that processes each channel independently to reduce computational complexity, and a Pointwise Convolution that recombines the channel information to extract more complex features.

To further refine the extracted features, the network incorporates a Channel Attention mechanism, which adaptively adjusts the importance of each channel. This mechanism utilizes global average pooling and max pooling to generate compact feature representations for each channel, which are then passed through two fully connected (FC) layers. These layers transform the pooled features, and finally, a Sigmoid function is applied to generate the channel attention weights, ensuring that the network focuses on the most relevant features for the prediction task. Finally, a second convolutional layer processes the attention-weighted feature maps, producing the final prediction output. Through this carefully designed architecture, the CFNN achieves a balance between computational efficiency and high predictive accuracy, making it well-suited for applications involving complex scientific data.

It is worth mentioning that the CFNN is trained using normalized original values, rather than prequantized values. After cross-field prediction, prequantization is applied. This means that for the same field, the same model can be used under different error bounds, and it still achieves good results. This approach saves a lot of training time and reduces storage cost.
\begin{figure}[]
    \centering
   \includegraphics[width=0.95\linewidth]{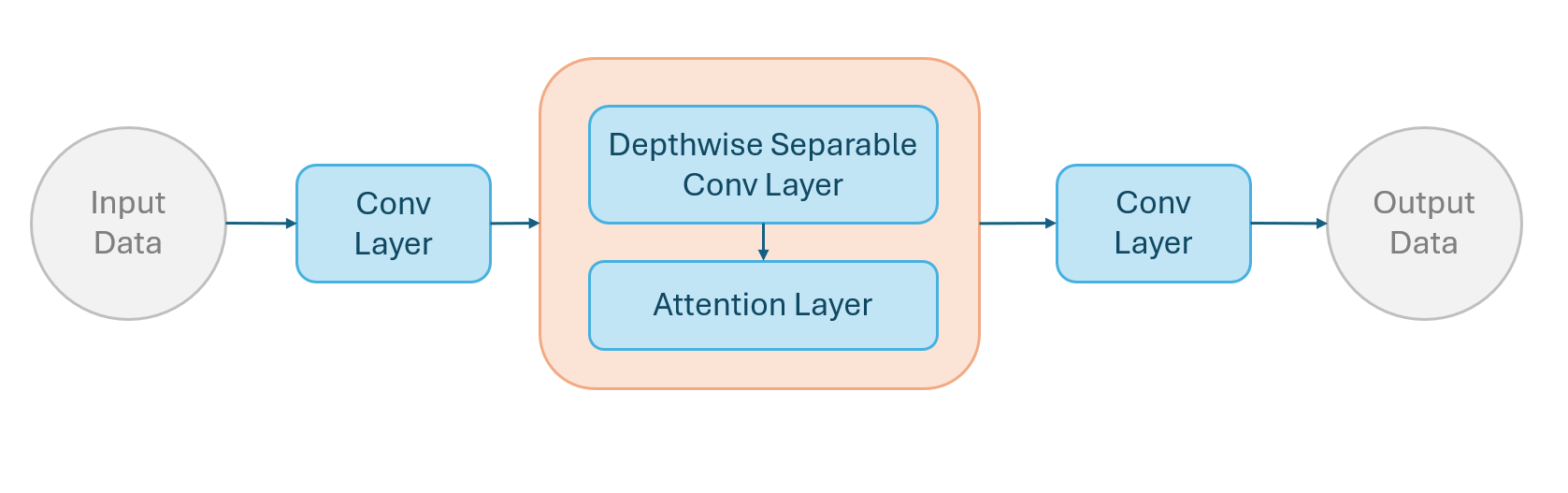}
   \vspace{-5mm}
    \caption{Overview of cross-field neural network (CFNN).}
    \label{fig:fig-cfnn}
\end{figure}
\subsubsection{Hybrid Prediction Model}
The design of the hybrid prediction model is significantly simpler compared to cross-field prediction. This simplification arises from the fact that, during decompression, the network is still constrained by RAW dependencies. While we addressed the RAW dependency issue during compression through the use of dual quantization, the decompression process still requires decompressing and refilling each element sequentially~\cite{tian2020cusz}. As a result, the network cannot be overly complex, as this would substantially reduce throughput. Our solution employs a neural network to directly learn a set of linear combination parameters, aiming to achieve improved predictions through a weighted summation of multiple predicted values. Experimental results indicate that this hybrid prediction model can achieve notable improvements, and due to its minimal computational overhead, it does not significantly impact throughput.

\section{Evaluation}

In this section, we present the evaluation results of our proposed framework for enhancing the performance of prediction-based lossy compression.
We first provide details about the experimental setup, datasets, and baseline configurations.
Next, we assess the effectiveness of cross-field information extraction to demonstrate its potential benefits. We then evaluate the performance of each individual component within our compression design.
Finally, we conduct a comprehensive and detailed performance evaluation using datasets from real-world scientific applications.

\subsection{Experimental Setup}
\subsubsection{System Configuration and datasets}
The experiments are conducted on the HPC cluster at Temple University. 
Each node equipped with two Intel Xeon E5-2620v4 processors featuring 16 physical cores, and 4x NVIDIA Tesla V100 GPUs. Datasets used in our experiments are listed in Table \ref{table:dataset}.
\begin{table}[]
\centering
\caption{Details of tested datasets}
\scriptsize
\renewcommand{\arraystretch}{1.5}
\begin{tabular}{@{}ccc@{}}
\hline
Name      & Dims         & Description          \\ \hline
Scale     & 98x1200x1200 & Climate simulation   \\ 
CESM(2D)  & 1800x3600    & Climate simulation   \\ 
Hurricane & 100x500x500  & Weather simulation   \\ \hline
\end{tabular}
\label{table:dataset}
\end{table}
\subsubsection{Baseline}
Our proposed model is designed to extract cross-field information and enhance the performance of prediction-based lossy compressors.
While the underlying approach is general and can be extended to other contexts, this paper focuses specifically on improving the existing SZ3 framework.
The baseline we use in our experiments is SZ3 with the Lorenzo predictor.
Additionally, our proposed solution uses dual-quantization to eliminate the prediction dependency, the corresponding SZ3 is also modified to use dual-quantization, while all other steps remain unchanged.
This also ensures that our solution is compatible with most CUDA versions of SZ~\cite{tian2020cusz}, where dual-quantization is natively integrated.
Note that our solution hybridizes information extracted from both cross-field and local-field predictions. Consequently, using the Lorenzo predictor (one of the local-field predictions) and dual-quantization as baselines ensures a fair comparison. 
We expect similar performance improvements with other compression configurations, which we plan to explore in future work.

\subsection{Cross-field Prediction}
We propose the CFNN in our solution to efficiently and effectively extract cross-field information.
The CFNN prediction the first-order backward difference of the target field based on data from other fields without relying on any local information from the target field. 
As discussed in Section~\ref{sec:3.1}, this design choice is based on the fact that local information is typically more dense and efficient to extract using existing predictors, such as Lorenzo.
Therefore, it would be suboptimal to apply the same method for extracting information from both cross-field and local-field contexts.

\begin{figure}[]
    \centering
   \includegraphics[width=0.95\linewidth]{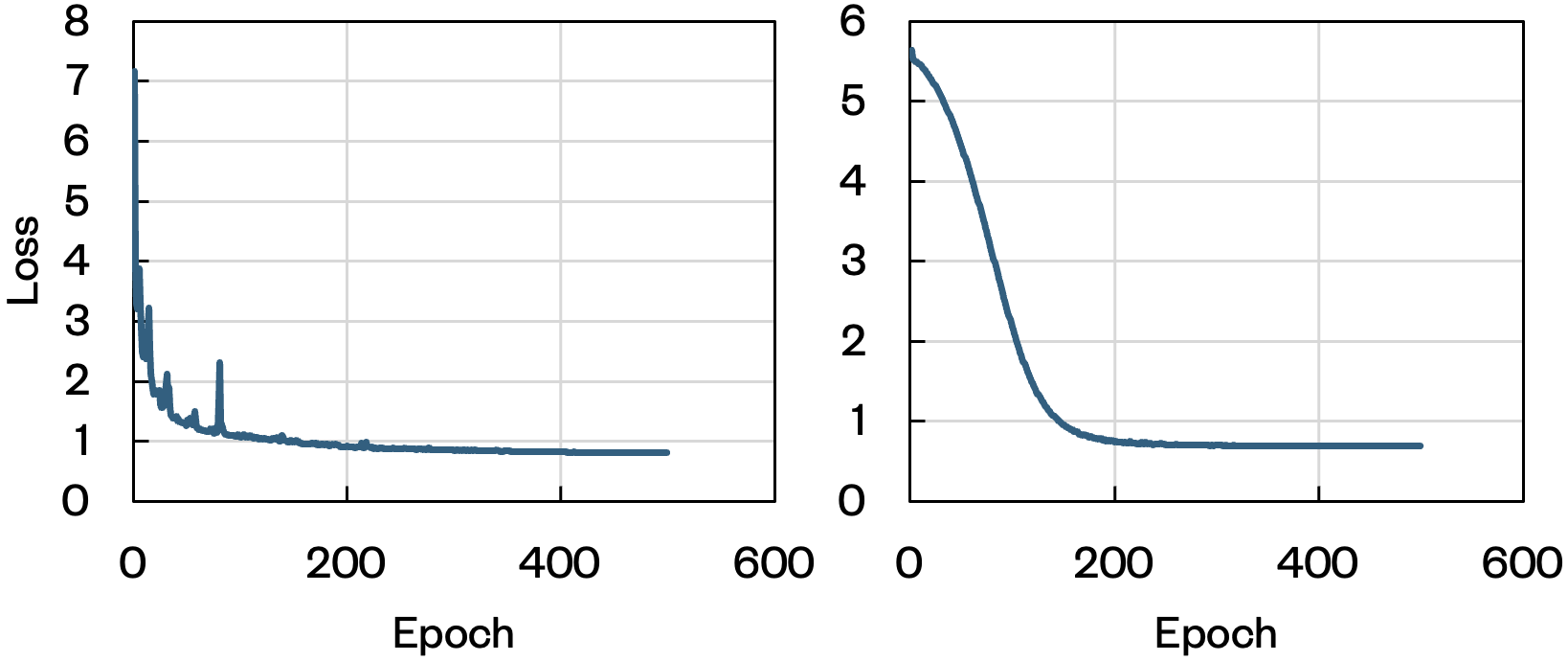}
    \caption{Training Loss vs Epoch during training. Left: CFNN model; Right: hybrid prediction model.}
    \label{fig:loss}
\end{figure}

Figure~\ref{fig:loss} shows the training loss during the model training for a 1e-3 relative error bound. The left plot illustrates the steady decrease in loss for the CFNN model across epochs, where the data has been normalized to the range of 0-300. The right plot displays the loss reduction for the hybrid prediction model, where the predicted data has already undergone prequantization, resulting in a value range of 0-1000. We employed Mean Squared Error (MSE) as the loss function in both models. The consistent decline in the training loss suggests that our approach is effectively extracting information from the anchor fields. Importantly, the absence of any abrupt changes or stagnation in the loss curves indicates that the models are learning appropriately without signs of overfitting or underfitting.

\begin{figure*}[]
    \centering
   \includegraphics[width=0.95\linewidth]{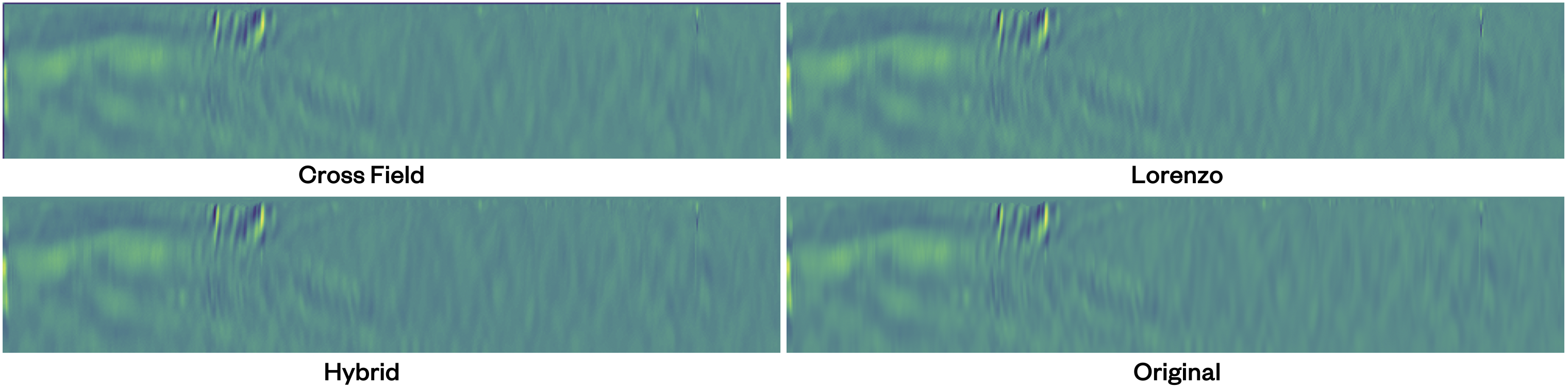}
    \caption{Visualization of the prediction accuracy through cross-field, lorenzo, and hybrid prediction. Showing the 50th slice of Wf, sliced along the second dimension. Dimension in 500$\times$100.}
    \label{fig:stepwise}
\end{figure*}

To illustrate the efficiency of cross-field prediction, we reconstruct the target field data using cross-field, Lorenzo, and hybrid predictions without applying error bound control (i.e., without using the second half of dual-quantization to encode prediction errors). In this setup, the cross-field reconstruction is based solely on the first-order backward difference predicted by CFNN. This approach ensures that the comparison fully reflects the prediction efficiency.

Figure~\ref{fig:stepwise} compares the reconstructed data obtained from cross-field prediction, local-field prediction (i.e., using the Lorenzo predictor), and our proposed hybrid prediction approach. 
This experiment was conducted on the Wf data field from the Hurricane dataset with an relative error bound of 1E-3. 
A prediction that closely matches the original data indicates higher prediction accuracy, leading to a more concentrated prediction error distribution and, consequently, a higher compression ratio.
For visualization, we extracted a slice along the second dimension, focusing on the 50th slice.
In this case, we use the Uf, Vf, and Pf fields to perform cross-field prediction on the Wf field. Here, Uf, Vf, and Wf represent zonal, meridional, and vertical (i.e., upward) wind speeds at 1000 hPa, respectively, with units in $ms^{-1}$.
Pf denotes pressure with units in $Pa$. The “f” suffix indicates that the data is in floating-point format.
Intuitively, these four data fields exhibit correlations that typically require significant domain knowledge from a scientist to fully extract. 
By leveraging our proposed cross-field prediction using CFNN, we observe that the reconstructed data closely resembles the original data when using cross-field information alone. 
It is important to note that this cross-field reconstruction does not rely on any local-field information.
Moreover, the relationship between the target field (Wf) and the anchor fields (Uf, Vf, and Pf) is inherently nonlinear, necessitating the use of a machine learning model--in this case, the CFNN--to effectively capture these correlations.

\begin{figure}[]
    \centering
    \includegraphics[width=0.75\linewidth]{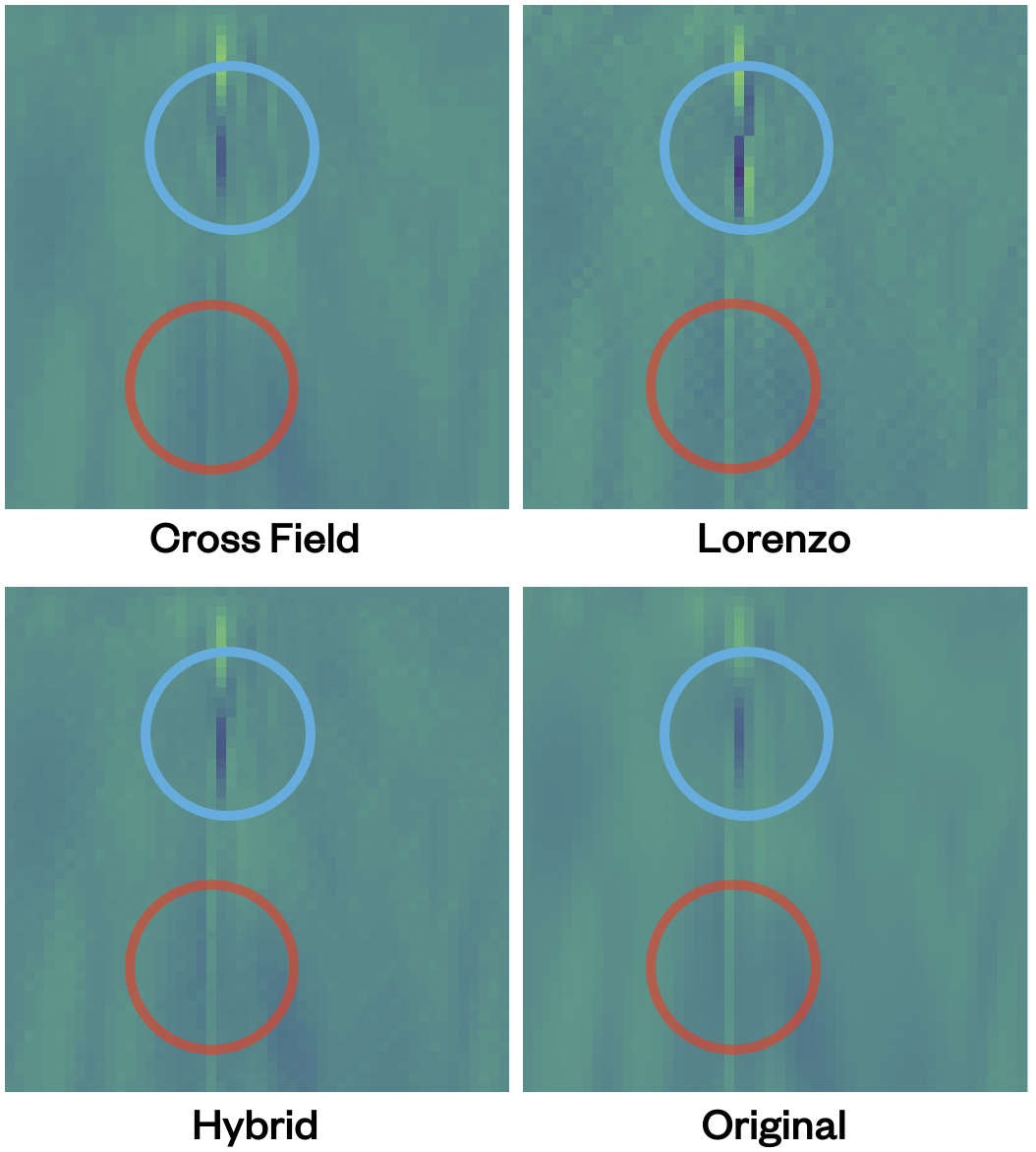}
    \caption{A zoom-in block in Figure \ref{fig:stepwise}. Dimension in 50$\times$50.}
    \label{fig:stepwise-detail}
\end{figure}

Figure~\ref{fig:stepwise-detail} shows a detailed comparison of a zoomed-in region from Figure~\ref{fig:stepwise}.
In the area highlighted by the blue circle, it is evident that the Lorenzo prediction exhibits noticeable artifacts, with blotchy color patterns that differ significantly from the smooth gradient of the original image.
In contrast, the cross-field prediction demonstrates higher precision in this region.
However, in the area marked by the red circle, the cross-field prediction shows less accuracy in predicting the bright color region in the middle, resulting in a lack of detail.
In this case, the Lorenzo prediction provides higher precision.
On the other hand, the Hybrid prediction avoids these artifacts and better preserves details in both regions.
This results in the highest overall prediction accuracy and, consequently, a higher overall compression ratio with the same error bound.

\definecolor{mygreen0}{RGB}{150, 150, 255}
\definecolor{mygreen1}{RGB}{50, 50, 205}
\definecolor{mygreen2}{RGB}{0, 0, 105}
\definecolor{myred}{RGB}{203, 107, 149}

\begin{table*}[]
\centering
\caption{Compression ratio of SCALE and Hurricane under different error bound and difference compression algorithm.}
\scriptsize
\renewcommand{\arraystretch}{1.3}
\vspace{-2mm}
\resizebox{.9\linewidth}{!}{
\begin{tabular}{|ccc|ccccc|}
\hline
\multicolumn{3}{|c|}{Error Bound} &
  \multicolumn{1}{c|}{5E-3} &
  \multicolumn{1}{c|}{2E-3} &
  \multicolumn{1}{c|}{1E-3} &
  \multicolumn{1}{c|}{5E-4} &
  \multicolumn{1}{c|}{2E-4} \\ \hline
\multicolumn{1}{|c|}{\multirow{6}{*}{Baseline}} &
  \multicolumn{1}{c|}{\multirow{2}{*}{SCALE}} &
  RH &
  \multicolumn{1}{c|}{/} &
  \multicolumn{1}{c|}{31.15} &
  \multicolumn{1}{c|}{25.75} &
  \multicolumn{1}{c|}{21.68} &
  16.14 \\ \cline{3-8} 
\multicolumn{1}{|c|}{} &
  \multicolumn{1}{c|}{} &
  W &
  \multicolumn{1}{c|}{/} &
  \multicolumn{1}{c|}{/} &
  \multicolumn{1}{c|}{27.48} &
  \multicolumn{1}{c|}{22.96} &
  19.29 \\ \cline{2-8} 
\multicolumn{1}{|c|}{} &
 \multicolumn{1}{c|}{\multirow{1}{*}{Hurricane}} &
  Wf &
  \multicolumn{1}{c|}{/} &
  \multicolumn{1}{c|}{25.13} &
  \multicolumn{1}{c|}{18.99} &
  \multicolumn{1}{c|}{15.98} &
  12.55 \\ \cline{2-8} 
\multicolumn{1}{|c|}{} &
  \multicolumn{1}{c|}{\multirow{3}{*}{CESM-ATM}} &
  CLDTOT &
  \multicolumn{1}{c|}{27.9} &
  \multicolumn{1}{c|}{20.72} &
  \multicolumn{1}{c|}{15.73} &
  \multicolumn{1}{c|}{11.65} &
  8.21 \\ \cline{3-8} 
\multicolumn{1}{|c|}{} &
  \multicolumn{1}{c|}{} &
  LWCF &
  \multicolumn{1}{c|}{/} &
  \multicolumn{1}{c|}{30.1} &
  \multicolumn{1}{c|}{23.64} &
  \multicolumn{1}{c|}{18.21} &
  12.2 \\ \cline{3-8} 
\multicolumn{1}{|c|}{} &
  \multicolumn{1}{c|}{} &
  FLUT &
 \multicolumn{1}{c|}{/} &
  \multicolumn{1}{c|}{/} &
  \multicolumn{1}{c|}{26.04} &
  \multicolumn{1}{c|}{20.68} &
  14.33 \\ \hline
\multicolumn{1}{|c|}{\multirow{6}{*}{Ours}} &
  \multicolumn{1}{c|}{\multirow{2}{*}{SCALE}} &
  RH &
  \multicolumn{1}{c|}{/} &
  \multicolumn{1}{c|}{32.44\textcolor{mygreen0}{(+4.12\%)}} &
  \multicolumn{1}{c|}{26.72\textcolor{mygreen0}{(+3.76\%)}} &
  \multicolumn{1}{c|}{21.51\textcolor{myred}{(-0.79\%)}} &
  \multicolumn{1}{c|}{15.6\textcolor{myred}{(-5.99\%)}} \\ \cline{3-8} 
\multicolumn{1}{|c|}{} &
  \multicolumn{1}{c|}{} &
  W &
  \multicolumn{1}{c|}{/} &
  \multicolumn{1}{c|}{/} &
  \multicolumn{1}{c|}{27.73\textcolor{mygreen0}{(+0.91\%)}} &
  \multicolumn{1}{c|}{21.32\textcolor{myred}{(-7.15\%)}} &
  \multicolumn{1}{c|}{16.28\textcolor{myred}{(-15.8\%)}} \\ \cline{2-8} 
\multicolumn{1}{|c|}{} &
 \multicolumn{1}{c|}{\multirow{1}{*}{Hurricane}} &
  Wf &
 \multicolumn{1}{c|}{/} &
  \multicolumn{1}{c|}{26.03\textcolor{mygreen0}{(+3.56\%)}} &
  \multicolumn{1}{c|}{22.72\textcolor{mygreen1}{(+19.63\%)}} &
  \multicolumn{1}{c|}{18.66\textcolor{mygreen1}{(+16.81\%)}} &
  \multicolumn{1}{c|}{13.72\textcolor{mygreen0}{(+9.38\%)}} \\ \cline{2-8} 
\multicolumn{1}{|c|}{} &
  \multicolumn{1}{c|}{\multirow{3}{*}{CESM-ATM}} &
  CLDTOT &
  \multicolumn{1}{c|}{28.54\textcolor{mygreen0}{(+2.32\%)}} &
  \multicolumn{1}{c|}{21.81\textcolor{mygreen0}{(+5.26\%)}} &
  \multicolumn{1}{c|}{17.15\textcolor{mygreen0}{(+9.03\%)}} &
  \multicolumn{1}{c|}{12.51\textcolor{mygreen0}{(+7.37\%)}} &
  \multicolumn{1}{c|}{8.26\textcolor{mygreen0}{(+0.62\%)}} \\ \cline{3-8} 
\multicolumn{1}{|c|}{} &
  \multicolumn{1}{c|}{} &
  LWCF &
  \multicolumn{1}{c|}{/} &
  \multicolumn{1}{c|}{31.45\textcolor{mygreen0}{(+4.51\%)}} &
  \multicolumn{1}{c|}{24.29\textcolor{mygreen0}{(+2.78\%)}} &
  \multicolumn{1}{c|}{20.27\textcolor{mygreen1}{(+11.28\%)}} &
  \multicolumn{1}{c|}{14.79\textcolor{mygreen2}{(+21.25\%)}} \\ \cline{3-8} 
\multicolumn{1}{|c|}{} &
  \multicolumn{1}{c|}{} &
  FLUT &
  \multicolumn{1}{c|}{/} &
  \multicolumn{1}{c|}{/} &
  \multicolumn{1}{c|}{27.56\textcolor{mygreen0}{(+5.87\%)}} &
  \multicolumn{1}{c|}{23.49\textcolor{mygreen1}{(+13.59\%)}} &
  \multicolumn{1}{c|}{18.31\textcolor{mygreen2}{(+27.76\%)}} \\ \hline
\end{tabular}
}
\vspace{-1mm}
\label{tab:metric}
\end{table*}

As mentioned in Section~\ref{sec:3.0}, we propose a hybrid prediction model to adaptively aggregates the prediction results from cross-field and local-field prediction. Our hybrid prediction model operates by performing a weighted sum of the \(n+1\) prediction results. During training, we observed that the model does not disproportionately favor any single predictor. Even in cases where the hybrid prediction does not significantly improve the compression ratio, the distribution of weights suggests that the model effectively captures the information embedded in the different predictors. For instance, in the Wf48 field, the hybrid prediction model assigns the highest weight (67\%) to the difference predictor in the z-axis direction, while the Lorenzo predictor receives the second highest weight at 25\%. The remaining two directional difference predictors are assigned 4\% each. Conversely, in the LWCF field, the Lorenzo predictor holds the highest weight at 60\%, followed by the x-direction difference predictor at 37\%. These results indicate that our model is capable of learning from the diverse information provided by each predictor. Moreover, considering the physical context—such as Wf48 representing upward wind speed—the higher weight assigned to the z-axis difference predictor is somewhat interpretable, reflecting not just a numerical relationship but also a physical correlation.

\subsection{Overall Performance Improvement}

In this subsection, we evaluate the overall performance of our proposed solution that hybrid both cross-field and local-field prediction.
We conducted experiments on two 3D datasets and one 2D dataset, , SCALE, Hurricane, and CESM-ATM, respectively. 
The comparison between the baseline and our method under different error bounds is detailed in Table \ref{tab:metric}. 
The experiments focus on scenarios where the baseline compression bit rate is above 1 bit (i.e., when the compression ratio of existing solutions is lower than 32 for floating-point datasets). 
This focus is justified for two reasons: 
(1) Datasets and data fields with lower compression ratios are significantly more important than those with higher ratios, as they often represent performance bottlenecks in scientific applications; 
(2) we expect our solution to provide more enhancements in compression performance in low-ratio cases. In high-ratio cases, local prediction alone is likely sufficient to extract the necessary encoding information without requiring assistance from cross-field prediction; 
and (3) The size of the machine learning model remains constant across different error bounds during compression, which means that the model’s proportion relative to the total compressed data size increases in cases with higher compression ratios.
Note that since dual quantization does not introduce additional errors after pre-quantization, the PSNR and other quality-related metrics for both methods are exactly the same. Therefore, we only report the compression rates in the table. 
The selection of anchor fields used as input for cross-field prediction is guided by basic physical principles (e.g., using wind speed to predict pressure), as detailed in Table~\ref{table:params}. 
It is one of our future work to develop a solution capable of automatically selecting anchor fields for a given dataset.
For example, we anticipate methods like transfer learning~\cite{xu22icml} could be used to determine the best prediction combinations. 
The preliminary selection of the current combinations of anchor fields and target fields has yielded varied results in the experiments.
\begin{table}[]
\caption{Experiment Configuration}
\scriptsize
\renewcommand{\arraystretch}{1.5}
\begin{tabular}{@{}c|c|c|c|c@{}}
\hline
Dataset & Target Field & Anchor Fields & \multicolumn{1}{c|}{\begin{tabular}[c]{@{}c@{}}Model Size\\ CFNN\end{tabular}} & \multicolumn{1}{c}{\begin{tabular}[c]{@{}c@{}}Model Size\\ Hybrid\end{tabular}} \\ \hline
\multicolumn{1}{c|}{\multirow{2}{*}{SCALE}} & \multicolumn{1}{c|}{\multirow{1}{*}{RH}} & T,QV,PRES  & \multicolumn{1}{c|}{32871} & \multicolumn{1}{c}{5}  \\ \cline{2-5}
\multicolumn{1}{c|}{} & \multicolumn{1}{c|}{\multirow{1}{*}{W}} & U,V,PRES  & \multicolumn{1}{c|}{32871} & \multicolumn{1}{c}{5}  \\ \hline
\multicolumn{1}{c|}{\multirow{1}{*}{Hurricane}} & \multicolumn{1}{c|}{\multirow{1}{*}{Wf}} & Uf,Vf,Pf  & \multicolumn{1}{c|}{32871} & \multicolumn{1}{c}{5}  \\ \hline
\multicolumn{1}{c|}{\multirow{6}{*}{CESM-ATM}} & \multicolumn{1}{c|}{\multirow{3}{*}{CLDTOT}} & CLDLOW  & \multicolumn{1}{c|}{} &  \multicolumn{1}{c}{}  \\ 
\multicolumn{1}{c|}{} & \multicolumn{1}{c|}{} & CLDMED  & \multicolumn{1}{c|}{5270} & \multicolumn{1}{c}{4}  \\ 
\multicolumn{1}{c|}{} & \multicolumn{1}{c|}{} & CLDHGH  & \multicolumn{1}{c|}{} &  \multicolumn{1}{c}{}  \\ \cline{2-5}
\multicolumn{1}{c|}{} & \multicolumn{1}{c|}{\multirow{1}{*}{LWCF}} & FLUTC,FLNT  & \multicolumn{1}{c|}{4470} & \multicolumn{1}{c}{4}  \\ \cline{2-5}
\multicolumn{1}{c|}{} & \multicolumn{1}{c|}{\multirow{2}{*}{FLUT}} & FLNT,FLNTC  & \multicolumn{1}{c|}{6070} &  \multicolumn{1}{c}{4}  \\ 
\multicolumn{1}{c|}{} & \multicolumn{1}{c|}{} & FLUTC,LWCF  & \multicolumn{1}{c|}{} & \multicolumn{1}{c}{}  \\ \hline
\end{tabular}
\label{table:params}
\end{table}

\begin{figure*}[]\centering
\begin{subfigure}{0.32\linewidth}\centering
    \includegraphics[width=1\linewidth]{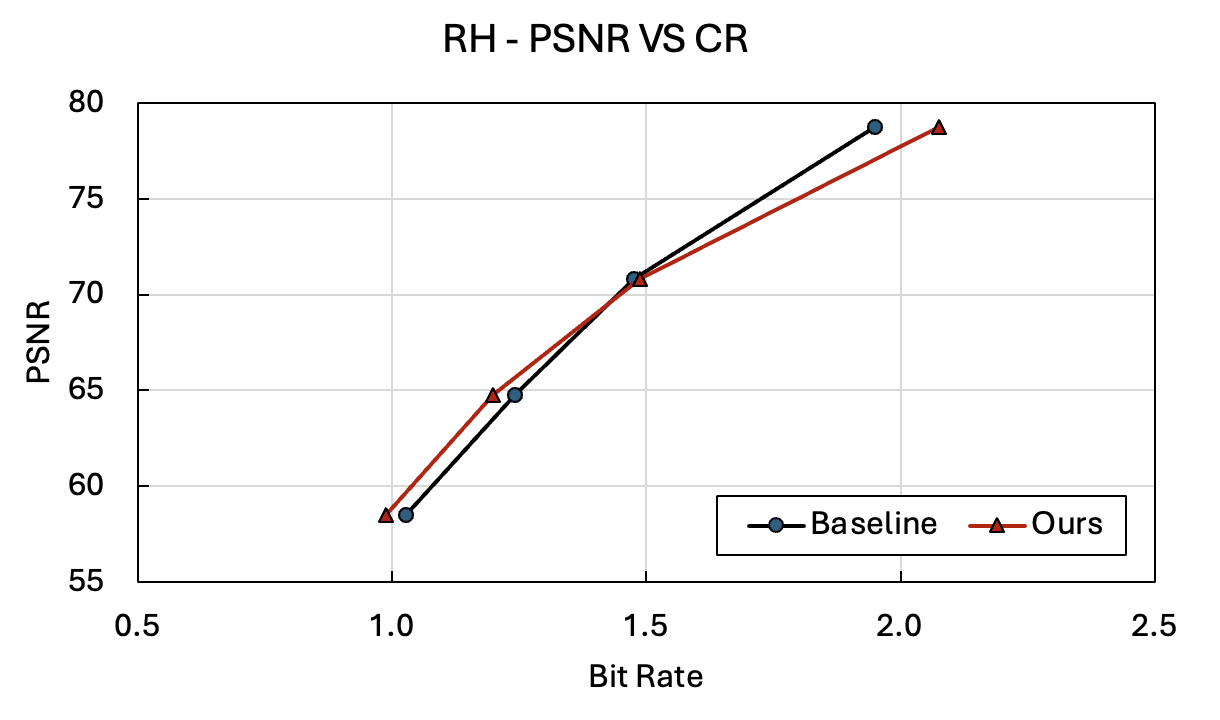}
	\caption{\footnotesize SCALE-RH}\label{subfig:rd-1}
\end{subfigure}
\begin{subfigure}{0.32\linewidth}\centering
    \includegraphics[width=1\linewidth]{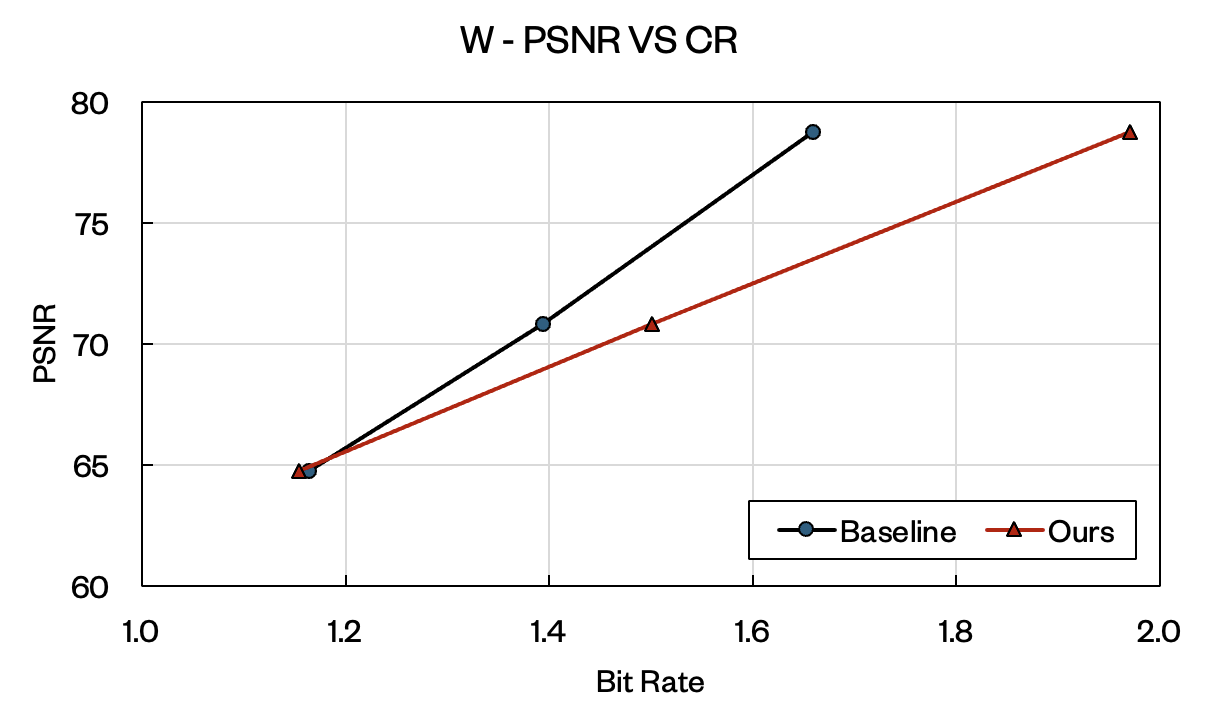}
	\caption{\footnotesize Scale-W}\label{subfig:rd-2}
\end{subfigure}
\begin{subfigure}{0.32\linewidth}\centering
    \includegraphics[width=1\linewidth]{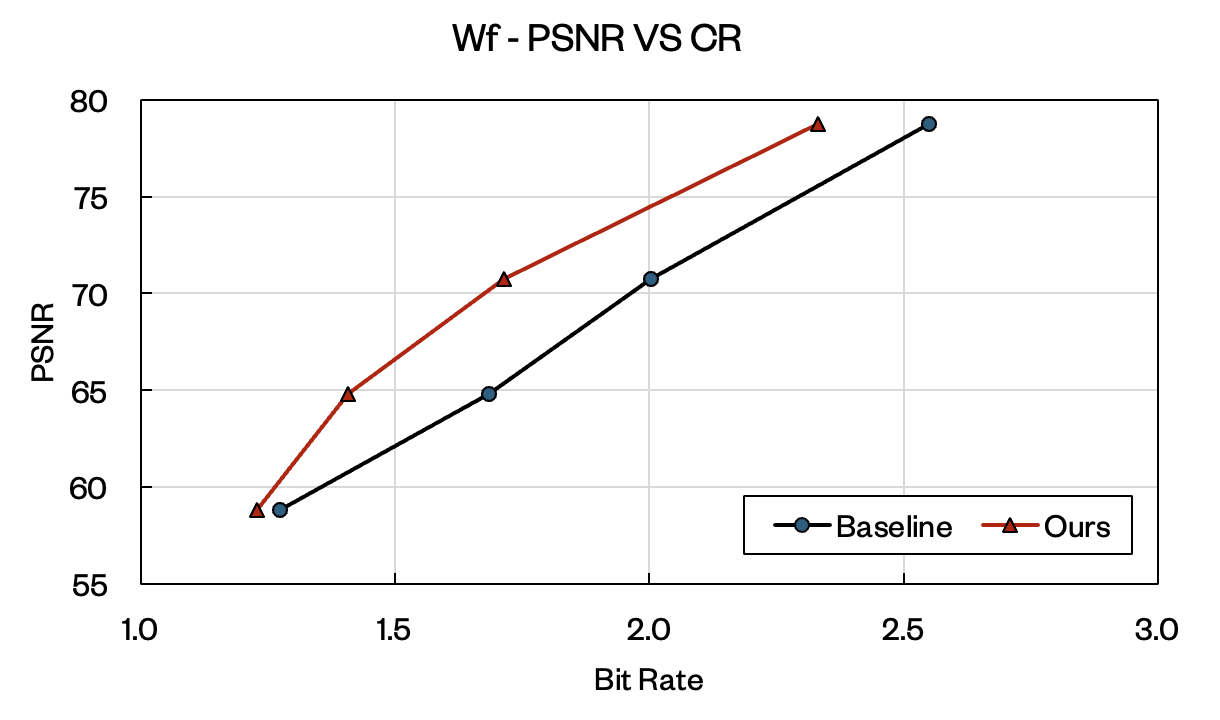}
	\caption{\footnotesize Hurricane-Wf}\label{subfig:rd-3}
\end{subfigure}
\begin{subfigure}{0.32\linewidth}\centering
    \includegraphics[width=1\linewidth]{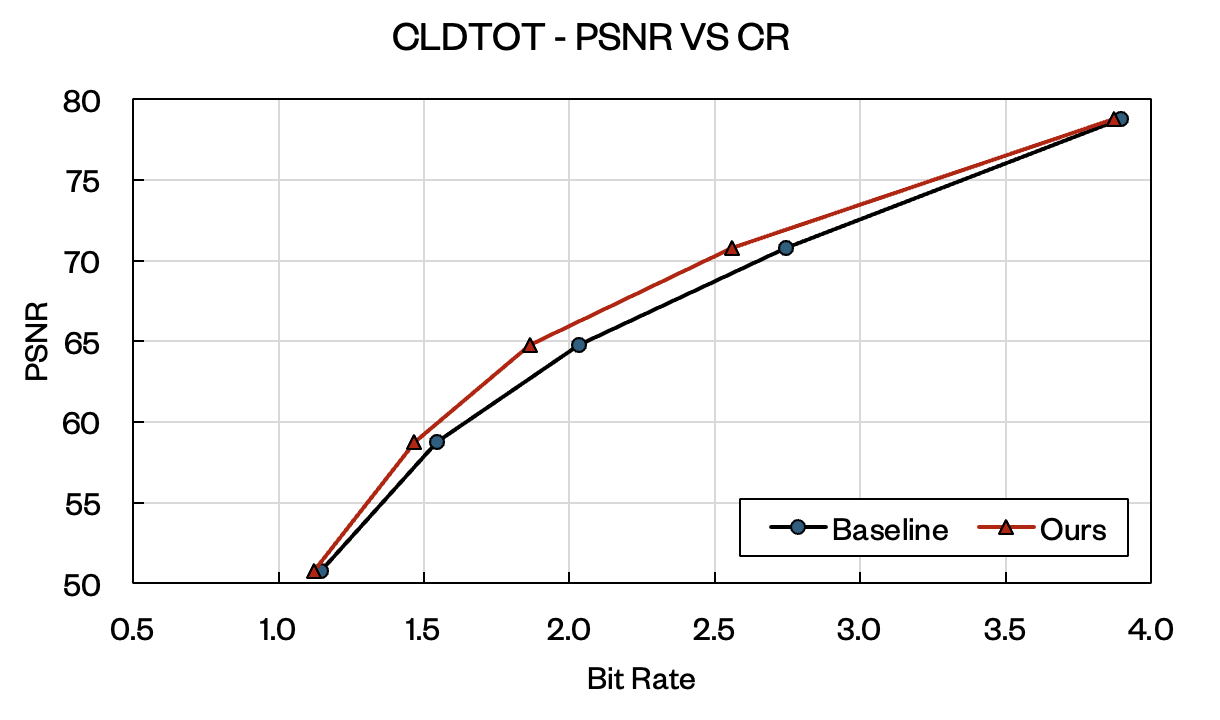}
	\caption{\footnotesize CESM-CLDTOT}\label{subfig:rd-4}
\end{subfigure}
\begin{subfigure}{0.32\linewidth}\centering
    \includegraphics[width=1\linewidth]{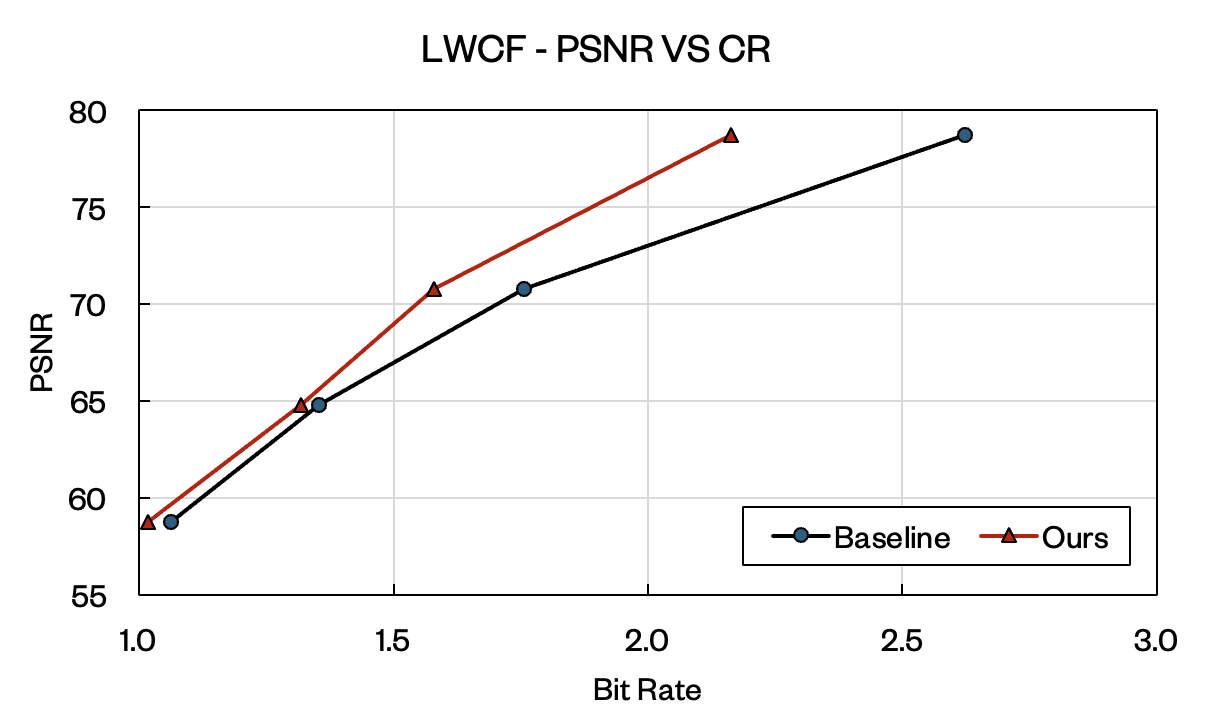}
	\caption{\footnotesize CESM-LWCF}\label{subfig:rd-5}
\end{subfigure}
\begin{subfigure}{0.32\linewidth}\centering
    \includegraphics[width=1\linewidth]{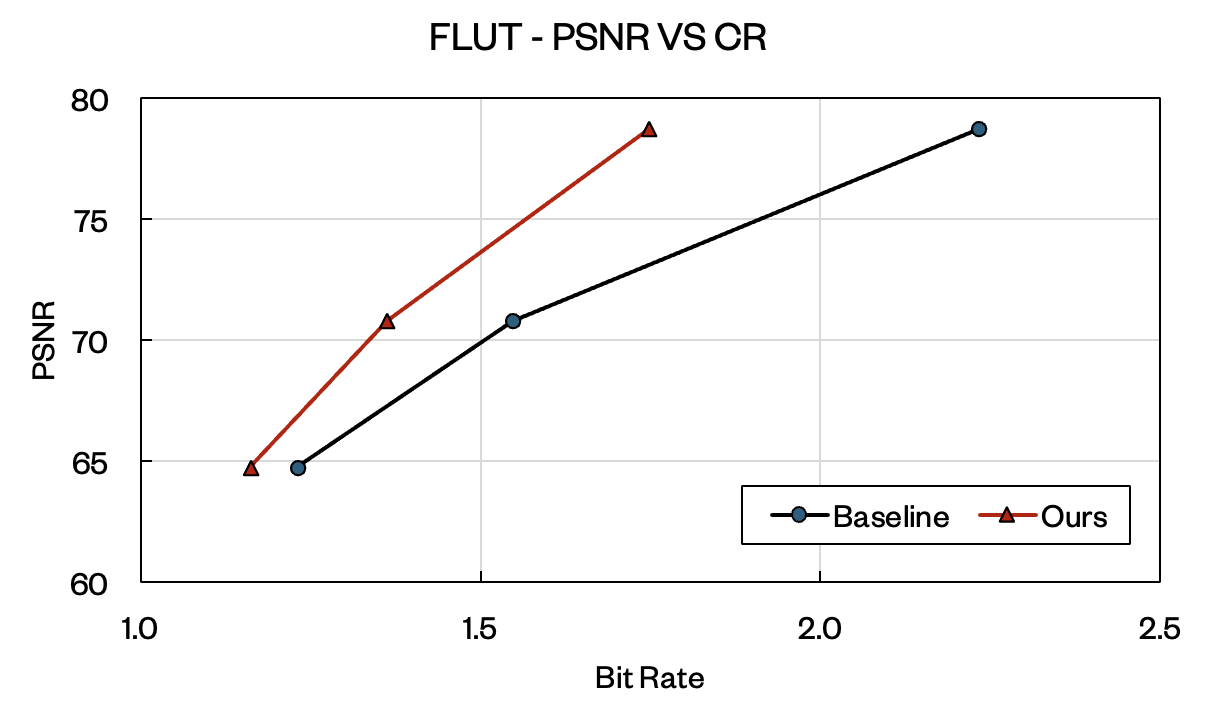}
	\caption{\footnotesize CESM-FLUT}\label{subfig:rd-6}
\end{subfigure}
\caption{Rate-distortion comparison between our solution and the baseline.}
\label{fig:overall-scale}
\end{figure*}

Our evaluation shows that in certain fields, such as the W field in the SCALE dataset, the compression ratio improvement can reach up to 25\% while maintaining the same error bound as previous solutions.
Most fields exhibit improvements in the 10\% range, there are cases where the gains are more modest, sometimes even below 10\%. 
In a few instances, the compression ratio may slightly decrease. 
This minor decrease is primarily due to the additional storage required by the CFNN model, which involves over 20,000 float32 parameters, detailed in Table \ref{table:params}. 
Another factor contributing to the less-than-expected results is that our hybrid prediction model is currently immature, relying only on a weighted sum. 
This simplicity limits its ability to fully integrate the information from two predictions. 
Despite this, the experiments shows the potential of our solution with cross-field prediction. 
As we refine the model and improve its capability to merge information, we anticipate even greater compression ratio improvements in the future. 
Additionally, in scenarios with high compression ratio or very small datasets, the machine learning model's relative size may impact the overall compression ratio. 
Overall, the consistent improvements demonstrate the robustness and adaptability of our approach across various datasets and conditions.

In addition to changes in compression rates under the same error bound, improvements in data quality and visualization fidelity are also crucial. 
Figure \ref{fig:overall-scale} shows the rate-distortion comparison results across different datasets.
This indicate that our solution consistently enhances data quality across various error bound. 
Note that the performance improvement tends to be more significant at higher bit rates (i.e., lower compression ratios). 
In the worst-case scenarios where our method yields lower compression performance, the degradations are subtle and are likely to be tolerable.

Additionally, figure~\ref{fig:fig-distortion-compare} compares a specific region, highlighted by the red rectangle, extracted from the original CLDTOT field and two decompressed versions at the same compression ratio of 17x. As highlighted by the red circles, the distortion introduced by the baseline method is significantly more noticeable under this compression ratio, whereas our solution shows much less distortion.
\begin{figure}[]
    \centering
   \includegraphics[width=0.95\linewidth]{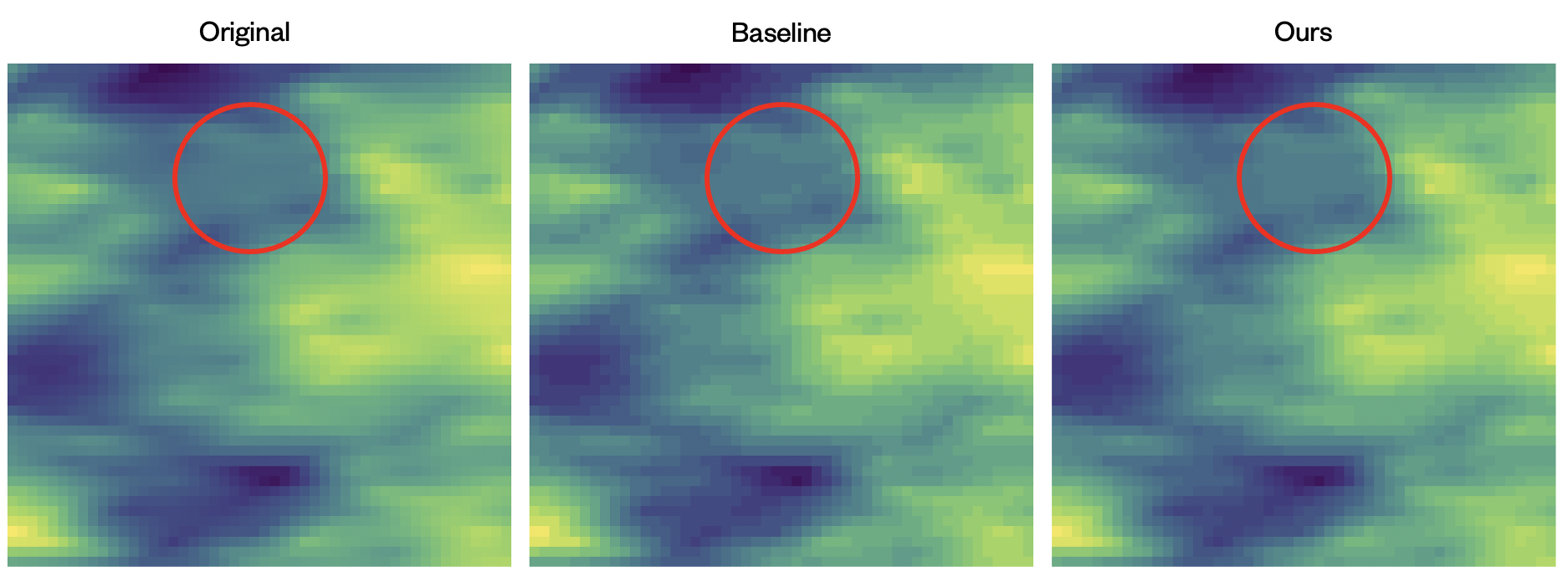}
    \caption{A zoom-in comparison of the original CLDTOT field in CESM with two decompressed versions at a 17x compression ratio. Dimension in 50$\times$50.}
    \label{fig:fig-distortion-compare}
\end{figure}

\section{Conclusion and Future Work}
\label{sec:conclusion}
In this paper, we introduce a novel cross-field prediction solution to enhance the performance of prediction-based lossy compressors. Our solution leverages CNNs to extract cross-field information, which is then combined with traditional predictors, effectively improving the overall compression ratio. Evaluations show that our solution can achieve compression ratio improvements of up to 27\% across multiple scientific datasets. 
Additionally, our solution preserves more data details and reduces artifacts compared to baseline methods.

In the future, we plan to optimize the structure of the CFNN to better extract cross-field information. Additionally, we will refine the hybrid prediction model to more effectively combine cross-field information with existing information. We will also explore the use of transfer learning to identify more suitable anchor fields for cross-field predictions. 

\section{Acknowledgement}
This research includes calculations carried out on HPC resources supported in part by the National Science Foundation through major research instrumentation grant number 1625061 and by the US Army Research Laboratory under contract number W911NF-16-2-0189. We also acknowledge the data resources provided on SDRBench~\cite{sdrbench,Zhao_2020}, which is maintained by Argonne National Laboratory.


\newpage
\bibliographystyle{IEEEtran}
\bibliography{refs}

\end{document}